\documentclass{article}

\PassOptionsToPackage{numbers,sort&compress}{natbib}
\usepackage[final]{neurips_2024}

\usepackage[utf8]{inputenc} % allow utf-8 input
\usepackage[T1]{fontenc}    % use 8-bit T1 fonts
\usepackage{hyperref}       % hyperlinks
\usepackage{url}            % simple URL typesetting
\usepackage{booktabs}       % professional-quality tables
\usepackage{amsfonts}       % blackboard math symbols
\usepackage{nicefrac}       % compact symbols for 1/2, etc.
\usepackage{microtype}      % microtypography
\usepackage{xcolor}         % colors 
\usepackage{graphicx}
\usepackage{multirow}
\usepackage{subcaption}
\usepackage{bm}
\usepackage{amsmath,amssymb,mathtools}
\title{Are LLMs good pragmatic speakers?}

\author{%
  Mingyue Jian \\ 
  University of Edinburgh\\
 \texttt{s2531301@ed.ac.uk} \\
 \And
  N. Siddharth \\ 
  University of Edinburgh \\
  \texttt{n.siddharth@ed.ac.uk} \\
}
\newcommand{\llm}{\texttt{LLM}}
\newcommand{\rsa}{\texttt{RSA}}
\newcommand{\pcc}{\texttt{PCC}}
\newcommand{\srcc}{\texttt{SRCC}}

\begin{document}

\maketitle

\begin{abstract}
  Large language models (\llm{}s) are trained on data assumed to include natural language pragmatics,
  but do they actually behave like pragmatic speakers?
  We attempt to answer this question using the Rational Speech Act (\rsa) framework \cite{frank2012pragmatic,goodman2016pragmatic}, which models pragmatic reasoning in human communication.
  Using the paradigm of a reference game constructed from the TUNA \cite{van2006building} corpus, we score candidate referential utterances in both a state-of-the-art \llm{} (Llama3-8B-Instruct) and in the \rsa\ model, comparing and contrasting these scores.
  Given that \rsa\ requires defining \emph{alternative} utterances and a truth-conditional \emph{meaning function}, we explore such comparison for different choices of each of these requirements.
  We find that while scores from the \llm\ have some positive correlation with those from \rsa, there isn't sufficient evidence to claim that it behaves like a pragmatic speaker.
  This initial study paves way for further targeted efforts exploring different models and settings, including human-subject evaluation, to see if \llm{}s truly can, or be made to, behave like pragmatic speakers.
  %
  % Trained with pragmatic data, are large language models (LLMs) capable of behaving like pragmatic speakers? 
  % % While most current studies focus on evaluating LLMs' ability to understand the implicature of non-literal content, there has been limited investigation into their ability to generate such content. 
  % We present an evaluation for the communicative effectiveness of LLMs using the Rational Speech Act framework, which is a probabilistic method for understanding pragmatic reasoning in the context of human communication. We compare the probabilities assigned by the LLM (MiniCPM-Llama3-V 2.5) to certain utterances against the probabilities assigned by an RSA model. We find that the LLM excels in factual judgement task for natural language, however, it lacks the capacity to distinguish between different levels of pragmatic generation as humans do.
\end{abstract}

\vspace*{-\baselineskip}
\section{Introduction}
% research question & gap
With the emergence of large language models (\llm{}s) \cite{gpt3,gpt4,chowdhery2022palm, lewis2019bart, lieber2021jurassic, taylor2022galactica, touvron2023llama}, a key question arises: can these models, trained on data presumed to include natural language pragmatics, exhibit pragmatic reasoning akin to humans? While \llm{}s have demonstrated formal linguistic competence (adherence to linguistic rules), their functional competence in pragmatic language use remains uncertain and warrants further investigation \cite{mahowald2024dissociating}. 
% Understanding the extent to which \llm{}s can engage in pragmatic reasoning is critical for evaluating their reliability and may also provide insights into human pragmatics \cite{hu2022fine}. 
Although much research has focused on evaluating \llm{s}' pragmatic abilities as listeners, particularly their comprehension of non-literal language \cite{hu2022fine,louis2020d, ruis2024goldilocks, sravanthi2024pub, nematzadeh2018evaluating}, less attention has been given to their pragmatic capabilities as speakers. Specifically, it remains unclear whether \llm{}s can effectively use context to generate non-literal utterances; i.e., being \emph{informative} beyond being simply \emph{true}. Investigating this aspect is crucial for deepening our understanding of \llm{}s' generative processes and enhancing their reliability.

Humans are typically pragmatic agents in communication. Imagine a room with three pieces of furniture: a small red desk, a small yellow desk, and a large red chair. If a friend directs your attention to ``the red one'', you would first eliminate the yellow desk as a possibility. Given the remaining red objects, your reasoning would then likely discount the chair, on account of it being distinct in the room, and that if it were the intended referent, the simpler utterance would have been just ``the chair''. This finally leads you to the intended referent being the small red desk.
This recursive reasoning process that takes \emph{informativity} into account beyond simply being \emph{literally} true, encapsulates pragmatic communication.

A particularly influential model of pragmatic communication is the Rational Speech Act (\rsa) framework \cite{goodman2016pragmatic,degen2023rational} which quantitatively models theory of mind \cite{dennett1997mind, premack1978does}, formalising how speakers and listeners use context, shared knowledge, and probabilistic reasoning to communicate effectively. This framework operates language understanding as a recursive process, where both speakers and listeners in a conversation behave rationally to reason each other's intention. 

% Speakers choose their words to optimally convey their message, anticipating that listeners will interpret these words in the most likely context. Meanwhile, listeners use the context and their linguistic knowledge to infer the speakers' intended meaning. 

In the earlier example, when a friend refers to ``the red one'', you exclude the yellow desk as a literal listener, guided by a ``meaning function'' that checks alignment with the words. The pragmatic speaker, your friend, chooses words with the expectation that you, as the listener, will interpret them correctly. As a pragmatic listener, you refine your interpretation, using both the message and context to infer intent. This interaction between speaker and listener, each considering the other’s perspective, is key to effective communication, as modelled by the \rsa\ framework.

\citet{nguyen2023language} applies the \rsa\ framework to view RLHF-fine-tuned language models as bounded pragmatic speakers, where RLHF equips the \llm{} with a ``theory of mind'' listener model. This enables the \llm{} to anticipate listener interpretation when computing the distribution of the pragmatic speaker. However, this study does not examine the pragmatic generation ability of unmodified \llm{}.
\citet{carenini2023large} examines vanilla \llm{}s' pragmatic reasoning using \rsa\ to find that GPT-2 XL’s reasoning aligns with \rsa\ in a metaphor task structured as ``$X$ is $Y$”, with a restricted meaning and utterance space. However, this focuses on the \llm{} as a \emph{listener}; we instead propose using a reference game to compare \llm\ scores as a \emph{speaker} against \rsa\, using a natural language format aligned with \llm{} training data. Additionally, we explore how the alignment varies with different sources of alternative utterances and \rsa\ models that employ distinct meaning functions.

% contribution

% There are mainly three contributions of this study. Firstly, we propose a pipeline for the pragmatic generation evaluation for LLMs using the RSA framework at inference. The pipeline is model-agnostic, which could be applied to different models for comprehensive analysis. Secondly, we present two methods to construct the utterance space of the reference game of the RSA framework. Thirdly, we present two methods of constructing the meaning function within the RSA framework. Our research provides insights into the pragmatic generation capabilities of the MiniCPM-Llama3-V 2.5 model \cite{yao2024minicpm, touvron2023llama} and examines how its correlation with the RSA model varies depending on different types of utterances. Additionally, we analyse how the alignment is affected when the RSA model is applied with different meaning functions. 
Our results indicate that our vanilla \llm{} model (Llama3-8B-Instruct \cite{touvron2023llama}) has a positive correlation with the two \rsa\ models that employ different meaning functions in the context of the reference game task, the correlation is stronger when scoring the logic-constructed utterances, and when the \rsa\ model is with a rule-based meaning function. However, we do not see a clear alignment of the \llm{}'s pragmatic scoring with that from the \rsa\ models.

\section{Rational Speech Act Model (RSA)}
 The \rsa\ model iteratively refines a heterogeneous relation between alternative utterances $U$ and intended meanings $O$, such that the relations begin being purely literal, and is refined using pragmatic reasoning: $ U \times O \to [0,1]$. The framework begins with a literal listener $L_{lit}$:
 \begin{align}
    P_{L_{lit}}(o|u) &\propto M(u,o) \cdot P(o), \qquad \label{eq:1}
\end{align}
where, in the context of our reference game, each object is equally likely to be selected, resulting in a uniform prior $P(o)$. Thus, the literal listener’s interpretation relies entirely on the meaning function $M()$. The pragmatic speaker $P_{S_p}$ is constructed from a literal listener $L_{lit}$: 

\begin{align}
    P_{S_p}(u|o) &\propto e^{\alpha (\ln P_{L_{lit}}(o|u) - \ln |u|)} = \left(\frac{P_{L_{lit}}(o|u)}{|u|}\right)^\alpha. \label{eq:2}
\end{align}

Here, $|u|$ is the utterance length imposing a cost on longer productions. This cost function aligns with the maxims of a pragmatic speaker \cite{grice1975logic}, favouring the use of fewer attributes to convey the intended meaning within a controlled attribute space. Additionally, this approach ensures that comparisons between the \rsa\ models and \llm{} remain valid. The preset prompt to the \llm{} (Appendix \ref{app: topk-prompt}) instructs it to describe the object using as few words as possible, effectively serving as a cost function that indirectly penalises longer outputs. $\alpha$ is a parameter that scale the rational level of $S_p$. A higher $\alpha$ will sharpen the probability distribution and vice verse.

% $\alpha = 1$, allowing us to directly assess the inherent rationality of the language model without artificially amplifying or dampening its behaviour.  

\subsection{RSA model with different meaning functions}
We construct two \rsa\ models with different meaning functions (MFs) for further investigation.
They take the form \(U \to [0,1]\), indicating whether the utterance $u$ literally describes the object $o$. 

\textbf{Prompt-based MF:} This leverages the natural language understanding capabilities of \llm{}s for scoring. We use prompt engineering to generate numeric scores from the \llm, employing 3-shot prompting to guide the model with input-output examples that establish a fixed output template (Figure \ref{fig:3-shot} in Appendix \ref{app: mf_eval}). The prompt-based meaning function is defined as:
% $ M_{\text{p}}(u,o) = \frac{P(\text{Yes} \:| \:\text{LLM}(o,u))}{P(\text{Yes} \: \cup \:\text{No}\:| \:\text{LLM}(o,u))},$
\begin{equation}
    M_{\text{p}}(u,o) = \frac{P(\text{Yes} \:| \:\text{\llm}(o,u))}{P(\text{Yes} \: \cup \:\text{No}\:| \:\text{\llm}(o,u))},
\end{equation}
representing the probability of the model answering ``Yes''. The prompt template is refined through a process of trial and error (Appendix \ref{app: mf_eval}).

\textbf{Rule-based MF}: This is based on feature exclusion: an utterance $u$ that includes a feature contradicting those of the object $o$ does not describe $o$. For example, if $o$ is ``a large, grey chair facing forwards'', then the utterance $u$ as ``a green thing'' does not describe $o$ as the colour feature in $u$ contradicts that of $o$. We define the rule-based meaning function as:
\begin{equation}
    o = \{f_1, f_2, ... f_n\} \subsetneq F,
\end{equation}
\begin{equation}
    M_{\text{r}}(u,o) := \nexists w. (\exists f.f \in F \setminus o \land D(w,f) ) \land (w \in u),
\end{equation}
where $f_1,f_2,...$ are the specific features of the object $o$, $F$ is a full set of predefined features in the dataset, and $D$ is a relation containing $(w,f)$ iff word $w$ describes feature $f$. % This formula gives a Boolean result, which is mapped to $\{0,1\}$ with $1$ being True.

We evaluate the two MFs against human-labelled ground truth data and find that the rule-based function consistently identifies literal relationships for the logical-constructed sequences, and most of the top-k generated sequences (mean Acc = 99.9\%), while the prompt-based method occasionally falls short in both constructed sequences (mean Acc = 97.3\% for the 3-shot prompt-based method) (Appendix \ref{app: mf_eval}).

\section{Evaluation Pipeline}
We use a reference game \cite{rosenberg1964speakers, krauss1964changes} as the task, where a set of objects $O$ includes target object $o_t$. The speaker selects an utterance $u_t \in U$ to convey $o_t$ to the listener, who must identify it based on $O$ and utterance $u_t$. For this task, we employ the TUNA dataset (furniture domain) \cite{van2006building}, which organises each reference game around 7 objects with predefined attributes and features (Appendix \ref{app:tuna}).

We propose a pipeline for the evaluation (Appendix \ref{app:pipeline}), which is organised into three stages: first, constructing alternative utterance $U$ and meaning $O$ spaces within the context of the reference game; second, getting scores from the vanilla \llm{} and the \rsa\ models with different meaning functions for each alternative sequence; and finally, evaluating the output distribution across various metrics.

\subsection{Construction of the meaning and utterance space}
\textbf{Meaning space:} For each reference game, we map object attributes from the TUNA dataset into a noun phrase template to generate descriptions:
\texttt{a <SIZE>, <COLOUR> <TYPE> facing <ORIENTATION>}.
This results in the meaning space of a set of 7 object descriptions for each game.

\textbf{Utterance space:} In a reference game, the utterance space includes all alternative utterances within the restricted world that could describe any object in the meaning space. An optimal utterance space would encompass both literal and pragmatic expressions, enabling a thorough assessment of communicative effectiveness. However, even in a restricted setting, the construction of such an utterance space, accounting for the wide variety of sentence structures and connotations found in natural language, can be difficult. Previous research has predominantly concentrated on producing sentences that are pragmatic, rather than exploring the full spectrum of meaning generation \cite{white2020learning}. In pragmatic referring expression generation, this typically involves sampling from a learned model during inference \cite{white2020learning, andreas2016reasoning, monroe2017colors}. However, this method inherently produces only pragmatic sequences, as \llm{}s are presumed to be trained on pragmatic data.
We present two approaches for constructing the utterance space $U$. 

\textbf{Top-\(k\) alternatives:} This samples the top-\(k\) utterances from the \llm{} using beam search, generating pragmatic sequences with flexible phrasing. The \llm{} receives a prompt (Appendix \ref{app: topk-prompt}) containing the world context and a concise task description to identify the target object $o_t$, ensuring minimal instruction to test the \llm's inherent pragmatic reasoning. To maintain consistency, generation begins with both ``a'' and ``the'' to ensure a noun phrase format. We generate sentences then deduplicate semantically identical ones that differ only in minor details like punctuation. 

\textbf{Logical rule alternatives:} This approach constructs both pragmatic and literal utterances based on logical rules. This method is particularly effective in a text-based reference game setting, where a literal utterance includes all relevant features, while a pragmatic utterance may involve omitting some features. In particular, since the generated sequence should follow a noun-phrase format, the omission for the object feature would be rephrased as `thing'. The formalisation of the logical construction for the utterance space is as follows:
\begin{equation}
    F_* = \bigtimes_{A\in \text{Attributes}} (F_A \cup \{\varepsilon\}),
\end{equation}
\begin{equation}
    U_\text{logic}(O) = \{ u(\mathbf{f}) | \mathbf{f} \in F_* | \exists \: o \in O . \mathbf{f} \subseteq o \},
\end{equation}
where $F_*$ represents all possible combinations of features in the reference game setting, and $O$ is a set of objects in a particular game. $U_{\text{logic}}(O)$ is a set of possible utterances that describe objects in $O$. $u()$ is a function we define to transform the feature set $\mathbf{f}$ to an utterance. Appendix \ref{app: logic} displays the formalisation of the logical construction process with examples. 

\subsection{Getting scores from the two models}
% In the scoring phase, the aim is to examine the behaviour of the LLM as it generates a specific preset sequence, given the context and target object. 
We score alternatives in the \llm\ and in \rsa\ given the same reference game world.

\textbf{Vanilla \llm: } The logit probabilities, i.e., raw output values from the \llm{} before they are transformed into a probability distribution, directly indicate the model's preferences. These are used as scores to assess the \llm's behaviour:
\[
  p(u \mid O,o_t)
  = p(\bm{u} \mid \bm{c}(O,o_t))
  = \prod_{i=1}^Np(u_{i} \mid \bm{c}(O,o_t),\bm{u}_{1:i-1}),
  \label{eq:llm-score}
\]
% $p(u|O,o_t) = p(\bm{u}|\bm{c}(O,o_t)) = \prod_{i=1}^np(u_{i} |\bm{c}(O,o_t),\bm{u}_{1:i-1}),$
% \begin{equation}
%     p(u|O,o_t) = p(\bm{u}|\bm{c}(O,o_t)) = \prod_{i=1}^np(u_{i} |\bm{c}(O,o_t)\bm{u}_{1:i-1}),
% \end{equation}
where $\bm{c}(\cdot)$ is the prompt template (Appendix \ref{app: topk-prompt}) and \(N\) is token length. Scores for top-\(k\) alternatives are generated along with the sequence using beam search. For the logic-constructed alternatives, we compute the probability retroactively for each utterance. Many popular pre-trained \llm{}s excel in downstream tasks, but few are open-source and provide logit probabilities. For this project, we use the open-source Meta-Llama3-8B-Instruct model \cite{touvron2023llama, MetaLlama8BInstruct}. We access logit probabilities using the \texttt{python-llama-cpp} library.

\textbf{RSA models:} The scores are calculated using Eq.\ref{eq:1} and Eq.\ref{eq:2} with the two constructed meaning functions.
 
\section{Results}
\textbf{Data Overview:} We have 2,940 reference games, each with a set of utterances describing one object, and we generate 386,510 utterance instances in total. Of these, 88,310 are generated by top-\(k\) sampling, and 298,200 by logic-based rules.

\textbf{Evaluation metrics:} We test models using probability outputs from the vanilla \llm{} and two \rsa\ models: one with a prompt-based MF and the other with a rule-based MF. We assess the correlation between scores from the vanilla \llm{} and \rsa\ models using Pearson Correlation Coefficient (\pcc) for linear relationships and Spearman’s Rank Correlation Coefficient (\srcc) for ranking similarity.

\begin{figure}[!h]
  \centering
  % overall correlation
  \begin{subfigure}[t]{0.35\textwidth}
    \includegraphics[width=0.9\textwidth]{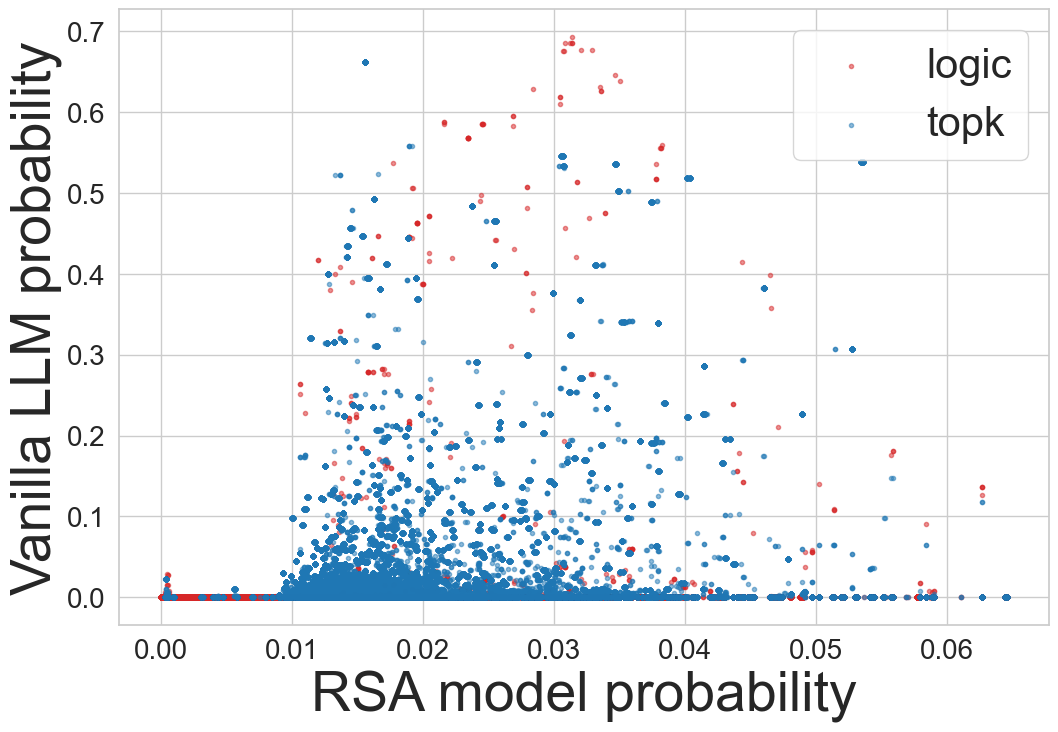}
    \includegraphics[width=0.9\textwidth]{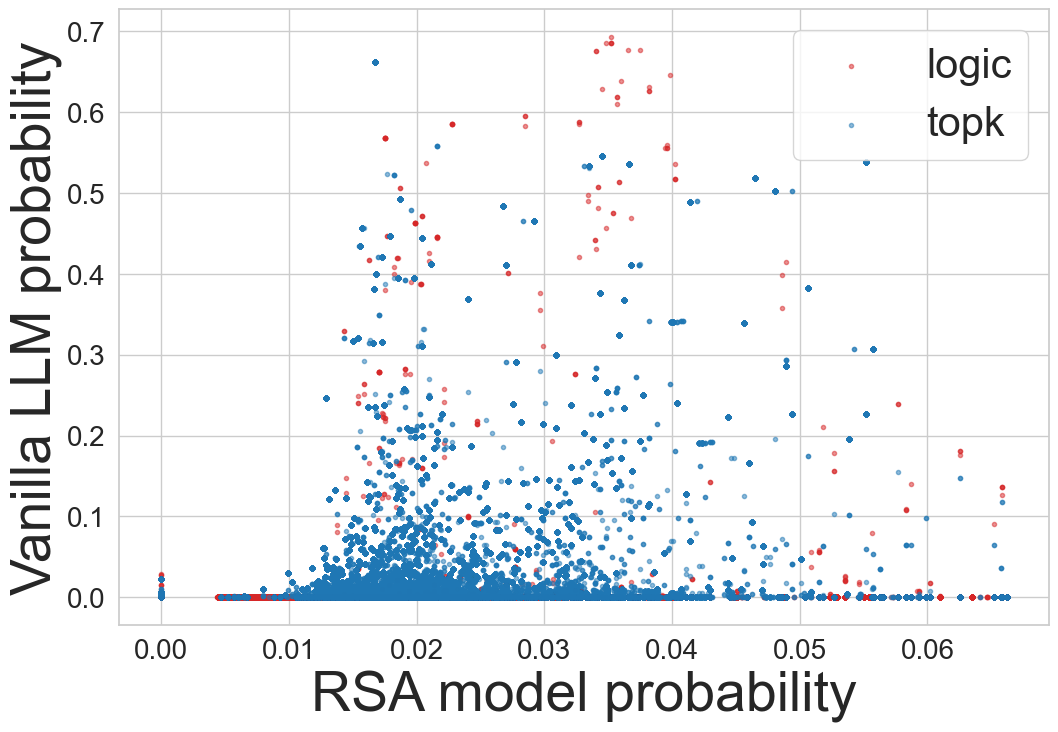}
    \caption{Overall correlation.}
    \label{fig:all}
  \end{subfigure}
  \;
  % per condition correlation histogram
  \begin{subfigure}[t]{0.63\textwidth}
    \begin{tabular}{@{}c@{}c@{}}
      \includegraphics[width=0.5\textwidth]{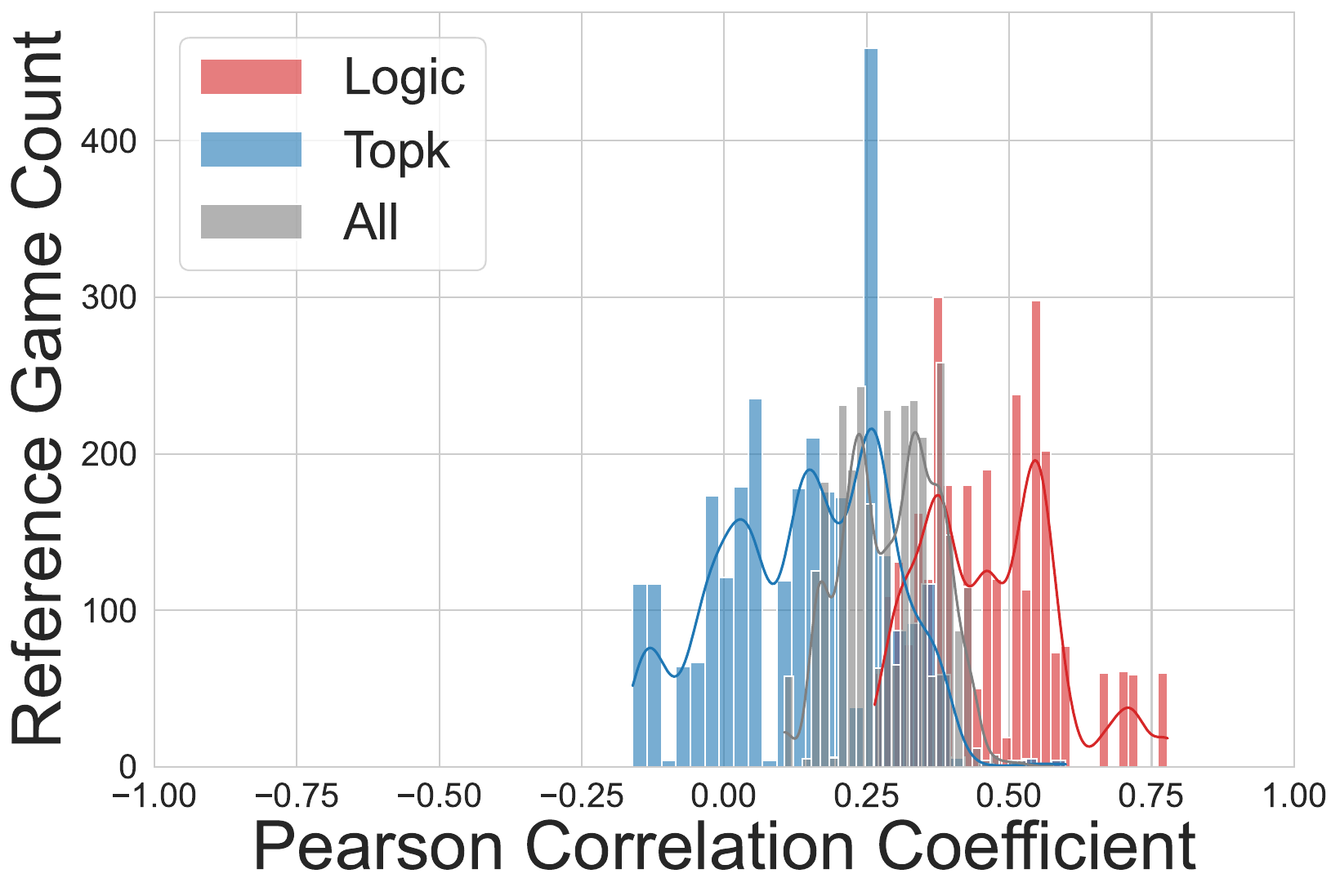}
      &\includegraphics[width=0.5\textwidth]{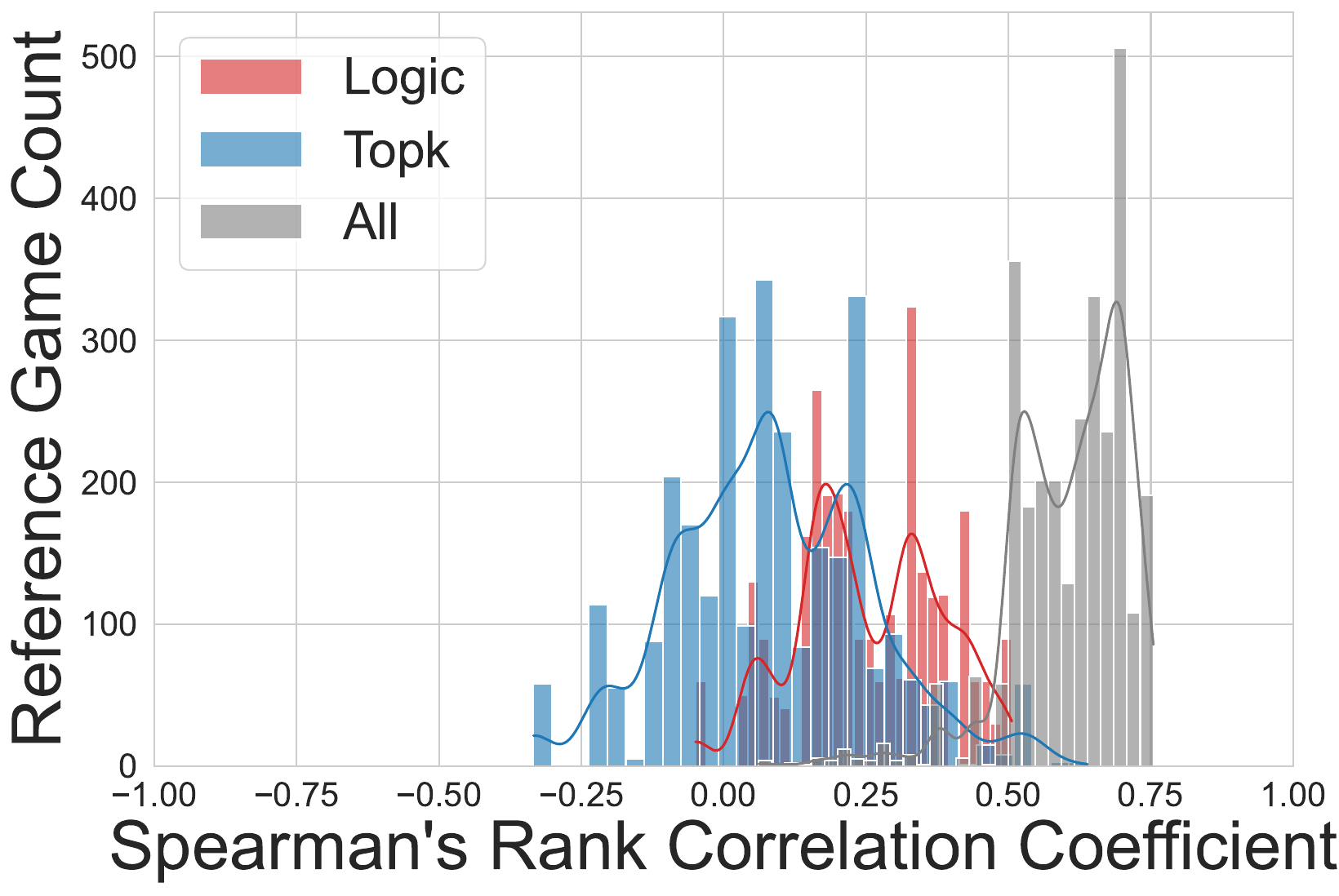} \\
      \includegraphics[width=0.5\textwidth]{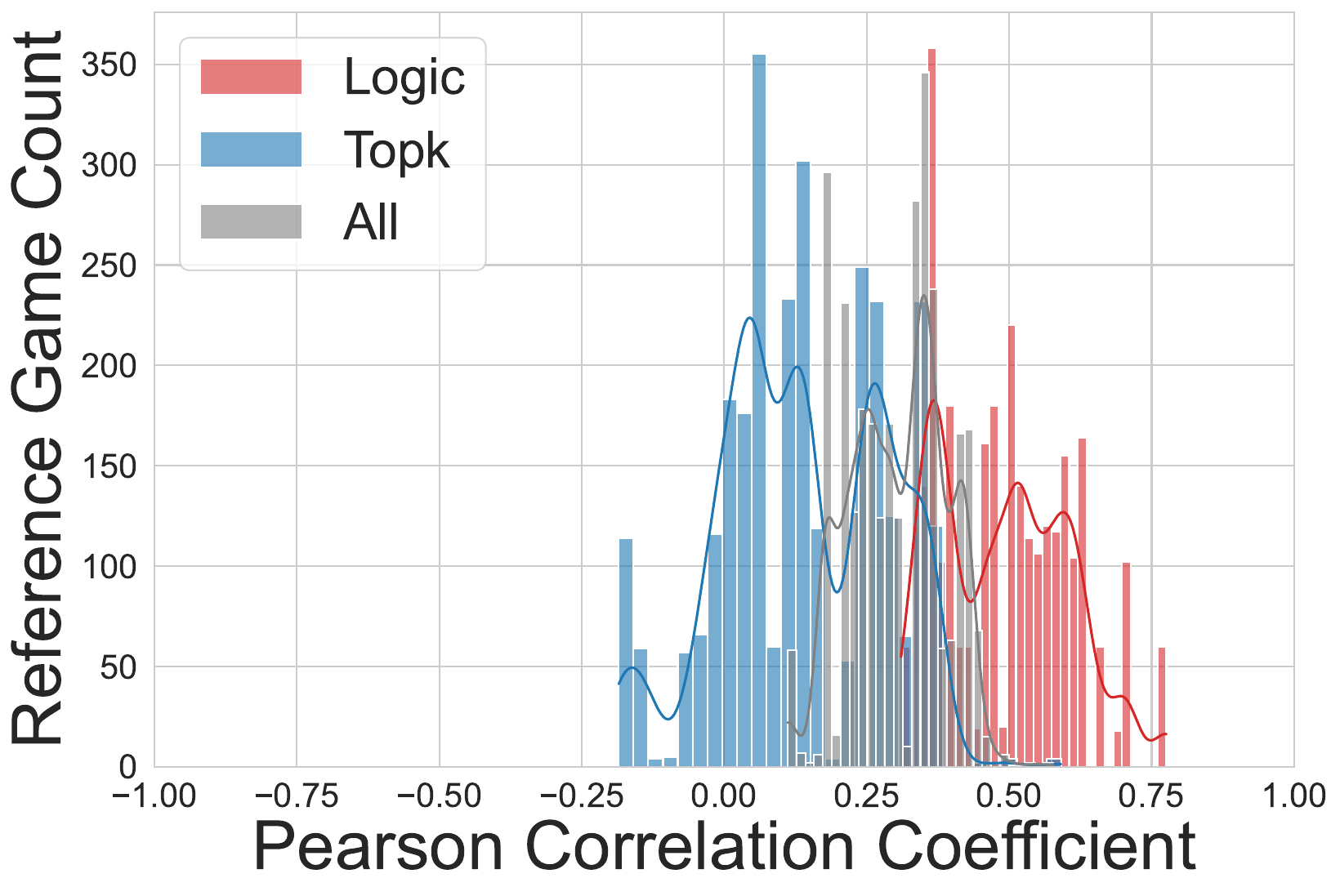}
      &\includegraphics[width=0.5\textwidth]{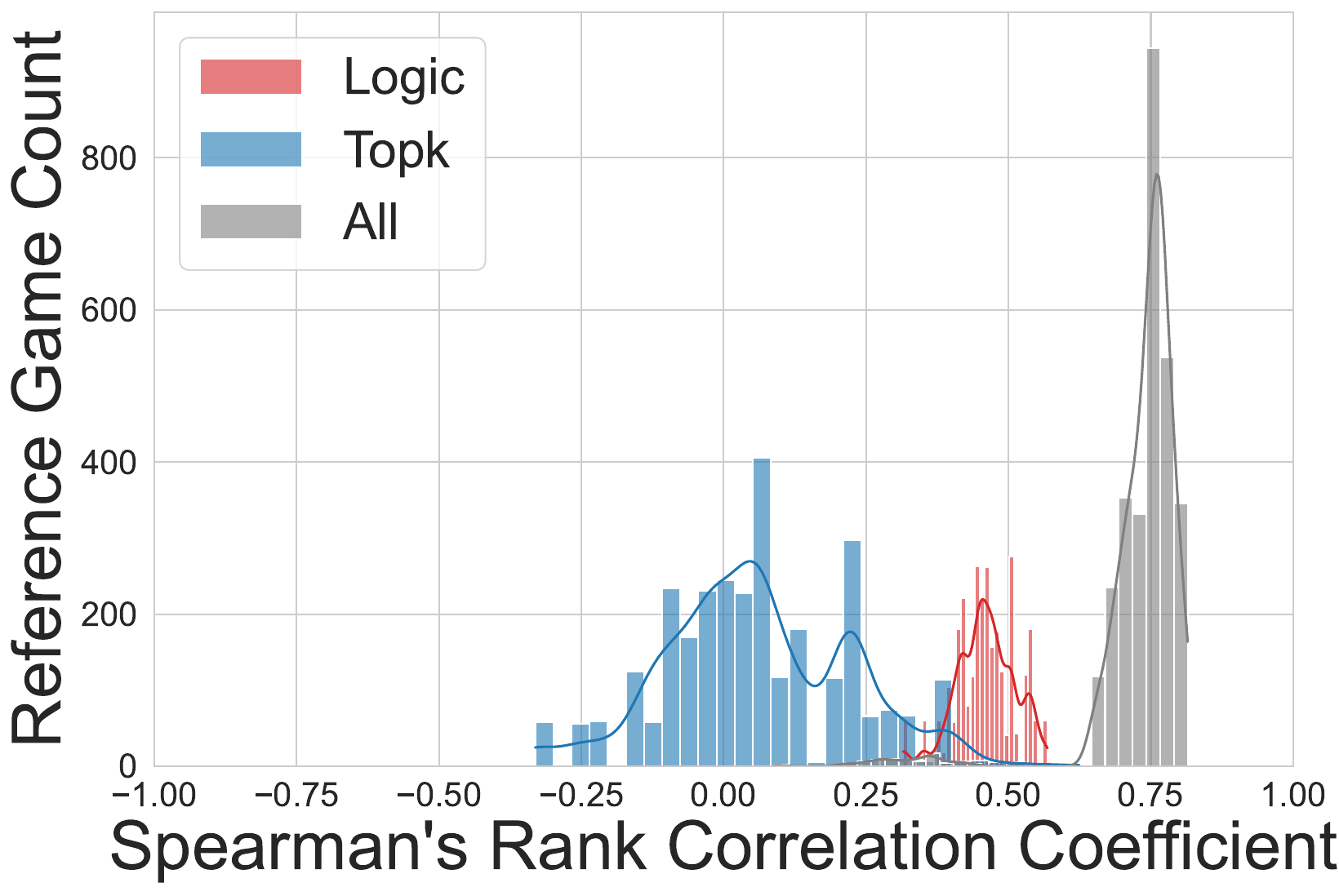}
    \end{tabular}
    \caption{
      Correlation histograms by condition with (l) \pcc\ and (r) \srcc.
    }
    \label{fig:breakup}
  \end{subfigure}
  \caption{
    Correlation analysis for scores in the \llm\ and \rsa\ models using (top) prompt-based MF and (bottom) rule-based MF.
    \textcolor{red}{Red} indicates logic-based alternatives, \textcolor{blue}{blue} indicates the top-\(k\) alternatives and \textcolor{gray}{grey} indicates all alternatives regardless of construction method.}
  \label{fig:experiments}
\end{figure}

\textbf{Experiment 1:} 
We analyse the overall correlation across all reference games by comparing the scoring of each utterance instance related to any object for both models. Figure \ref{fig:all} illustrates the correlation between the vanilla \llm{} and the \rsa\ models using different meaning functions. Both scatter plots show no clear linear relationship between the scores. The performance of the two meaning functions is comparable, as evidenced by the similar patterns observed across both plots. The \rsa\ model's scoring can be interpreted as a spectrum from incorrect to literal to pragmatic utterances. The vanilla \llm{} scoring reveals that literal utterances are favoured by the \llm{} more than pragmatic ones. Particularly notable is the right-hand tail of the graph, where many pragmatic utterances are either minimally acknowledged or largely overlooked by the \llm. This suggests that while the \llm{} is able to correctly make factual judgement, they are unlikely to rank the utterances pragmatically. The \rsa\ models in Figure \ref{fig:experiments} are configured with $\alpha = 1.0$. To further investigate the effect of $\alpha$ on correlation, we experiment with different values of $\alpha \in \{0.2, 0.6, 1.0, 1.4, 1.8, 3.0\}$. Correlation plots for each $\alpha$ are provided in Appendix \ref{app: alpha}. Our results indicate that varying $\alpha$ does not alter our primary finding: the scores from \llm{} do not exhibit strong alignment with those of the \rsa\ models, and the general distribution pattern remains consistent. Furthermore, we observe that increasing $\alpha$ sharpens the probability distribution, as reflected by the increase of the upper-bound on the x-axis.

\textbf{Experiment 2:} We then compare models by assessing their scoring preferences for each set of utterances describing each object, the result shows positive correlation on \pcc\ and \srcc\ scores all model evaluations (Appendix \ref{app: experiment1}). Figure \ref{fig:breakup} displays the histogram for the correlation of each experimented instance on both metrics. The vanilla \llm{} aligns more closely with the \rsa\ model using a rule-based MF (\pcc\ =0.303, \srcc\ =0.736) than with the one using a prompt-based function (\pcc\ =0.291, \srcc\ =0.606). We can see that the correlation between the \llm{} and \rsa\ models is stronger for the logic-constructed utterances than for the top-\(k\) ones, likely due to the predictability of the former. In contrast, top-\(k\) utterances, prone to hallucinations, show more variability, highlighting the challenges \llm{}s face in maintaining coherence and accuracy in less structured tasks. 

Interestingly, when evaluating performance on \srcc\, we find that correlation scores across the entire sequence space for each reference game set are positive regardless of meaning function method, with most exceeding 0.5. This suggests that the \llm{} can effectively distinguish top-$k$ sequences from logically constructed sequences and rank them correctly to some extent. However, it lacks the ability to accurately rank the top-$k$ sequences within the group, as indicated by \srcc\ scores ranging from -0.25 to 0.50 for both \srcc\ graphs.

% Interestingly, when checking the performance on \srcc, we find that when scores are calculated across the entire sequence space for each reference game set, the distribution of correlation scores divides into two distinct groups. This separation likely stems from differences in utterance construction. While top-\(k\) utterances usually score higher for accuracy, frequent hallucinations reduce their correlation with \rsa\, allowing logic-constructed sequences to occasionally score higher, resulting in the observed separation in rank correlation across all utterance types.

\section{Discussion}
Based on our experiments and analyses, we see no clear evidence to suggest that \llm{}s are good pragmatic speakers. When comparing its scoring with the \rsa\ models using different meaning functions, the \llm{} aligns more with the rule-based MF, especially for logic-constructed utterances. In our meaning function evaluation, the rule-based MF outperforms the prompt-based approach in factual judgement tasks, highlighting the \llm’s strength in structured reasoning.

While these results highlight the \llm’s pragmatic abilities in a controlled setting, their generalisability to everyday language remains uncertain. The structured nature of the reference games may not fully reflect the complexities of real-world communication. However, this research offers a framework for evaluating \llm{}s' pragmatic abilities and could be extended to more natural language use.

Future work should explore more diverse datasets to reflect a wider range of communication settings and natural language use. Testing on other \llm{}s, especially those with advanced pragmatic reasoning like \texttt{GPT} models trained on large datasets, would provide deeper insights into handling pragmatic tasks. Further research could also compare \llm{} alignment with the \rsa\ model when iterated multiple times, rather than a single interaction, and examine the effects of scaling parameters and cost functions on alignment.

\bibliography{refs}

\newpage
\appendix

\section{TUNA dataset attributes and features}\label{app:tuna}

\begin{table}[!h]
    \centering
    \begin{tabular}{@{}cc@{}}
    \toprule
         \textbf{Attribute} & \textbf{Possible features}  \\
    \midrule
         Type & chair, sofa, desk, fan \\
         Colour & blue, red, green, grey \\
         Size & large, small \\
         Orientation & left, right, front, back \\
    \bottomrule
    \end{tabular}
    \caption{Preset attributes and corresponding possible features for the `furniture' domain in the TUNA dataset.}
    \label{tab:TUNA}
\end{table}

\section{Prompt template for getting top-k sequences and the scores from \llm} \label{app: topk-prompt}
\begin{figure}[!h]
    \centering
    \includegraphics[width=0.5\linewidth]{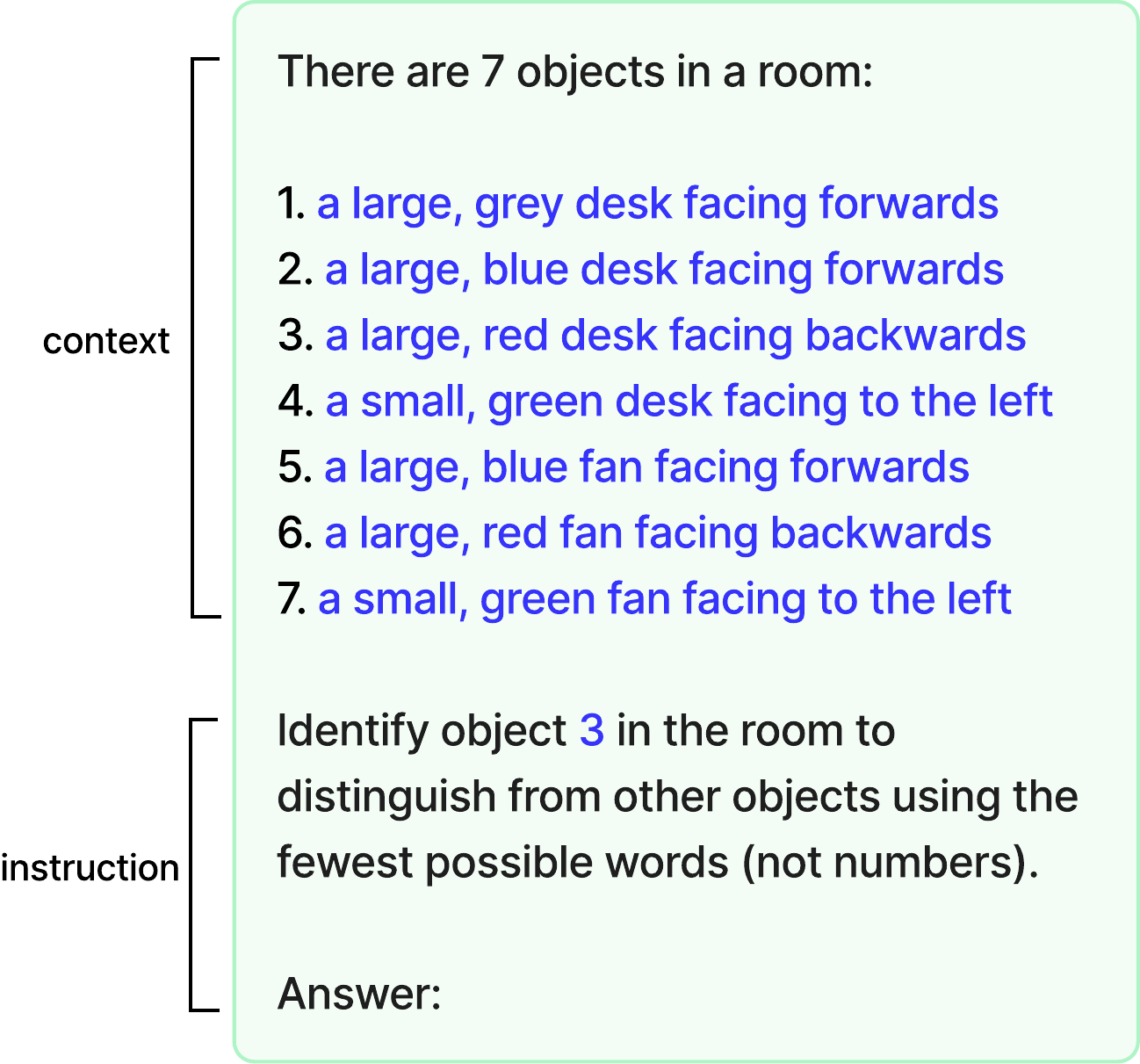}
    \caption{Example of the prompt used for generating top-k sequences with the \llm.
The blue text indicates variable elements specific to each reference game instance.}
    \label{fig:topk-prompt}
\end{figure}

\newpage

\section{Evaluation pipeline}\label{app:pipeline}
\begin{figure}[!h]
    \centering
    \includegraphics[width=0.95\textwidth]{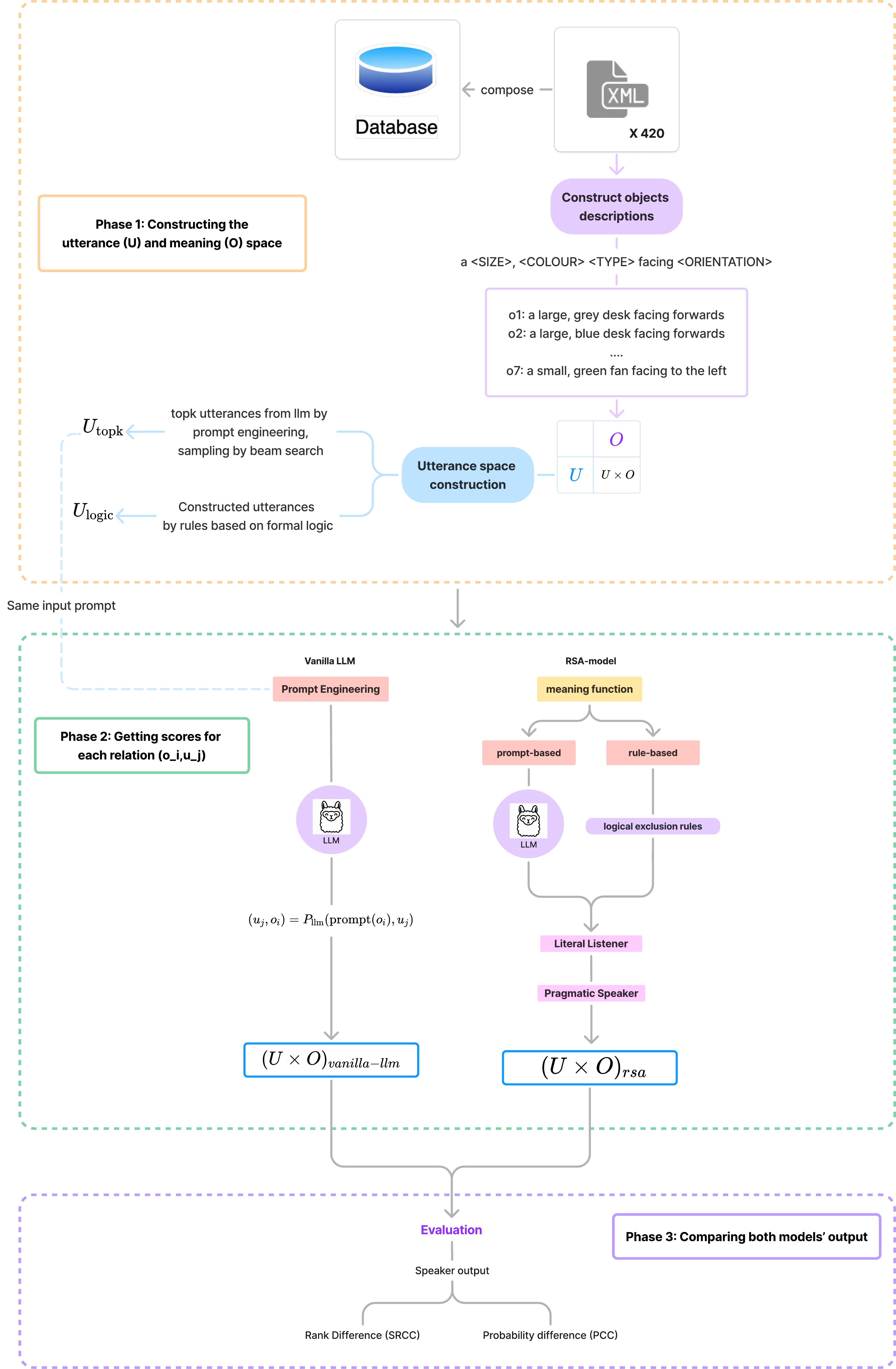}
    \caption{Project methodology pipeline}
    \label{fig:pipeline}
\end{figure}
\newpage
% Optionally include supplemental material (complete proofs, additional experiments and plots) in appendix.
% All such materials \textbf{SHOULD be included in the main submission.}

%%%%%%%%%%%%%%%%%%%%%%%%%%%%%%%%%%%%%%%%%%%%%%%%%%%%%%%%%%%%
\section{Logical construction process} \label{app: logic}
\begin{figure}[!h]
    \centering
    \includegraphics[width=1\linewidth]{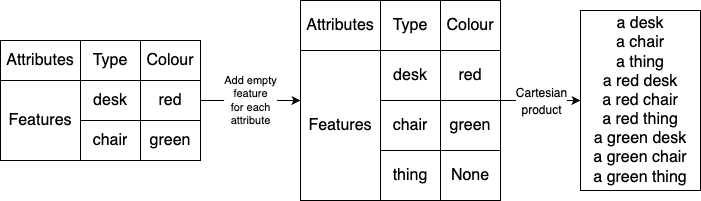}
    \caption{Example of logical construction process, given the attribute sets in the world.}
    \label{fig:logic}
\end{figure}

% The formalisation of the logical construction for the utterance space is as follows:
% \begin{equation}
%     F_* = \bigtimes_{A\in \text{Attributes}} (F_A \cup \{\varepsilon\}),
% \end{equation}
% \begin{equation}
%     U_\text{logic}(O) = \{ u(\mathbf{f}) | \mathbf{f} \in F_* | \exists \: o \in O . \mathbf{f} \subseteq o \},
% \end{equation}
% where $F_*$ represents all possible combinations of features in the reference game setting, and $O$ is a set of objects in a particular game. $U_{\text{logic}}(O)$ is a set of possible utterances that describe objects in $O$. $u()$ is a function we define to transform the feature set $\mathbf{f}$ to an utterance. 
Figure \ref{fig:logic} is a concrete example of a logical construction process.

\clearpage

\section{Results of the correlation scores of \llm{} and \rsa\ model using different $\alpha$ values} \label{app: alpha}

\begin{figure}[!ht]
    \centering

    % First pair: alpha = 0.2
    \begin{subfigure}{0.99\textwidth}
        \centering
        \begin{subfigure}{0.49\textwidth}
            \centering
            \includegraphics[width=\textwidth]{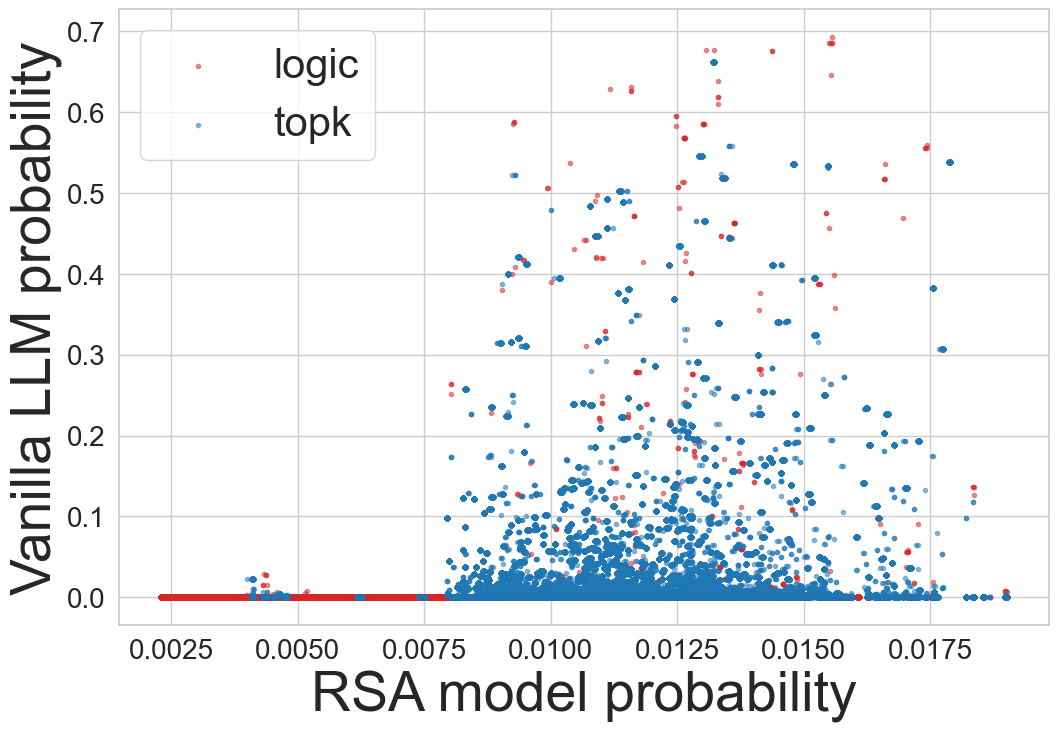}
            \caption{\pcc{}$_{topk}$ = 0.153, \pcc{}$_{logic}$ = 0.372, \pcc{}$_{all}$ = 0.259} 
        \end{subfigure}
        \hfill
        \begin{subfigure}{0.49\textwidth}
            \centering
            \includegraphics[width=\textwidth]{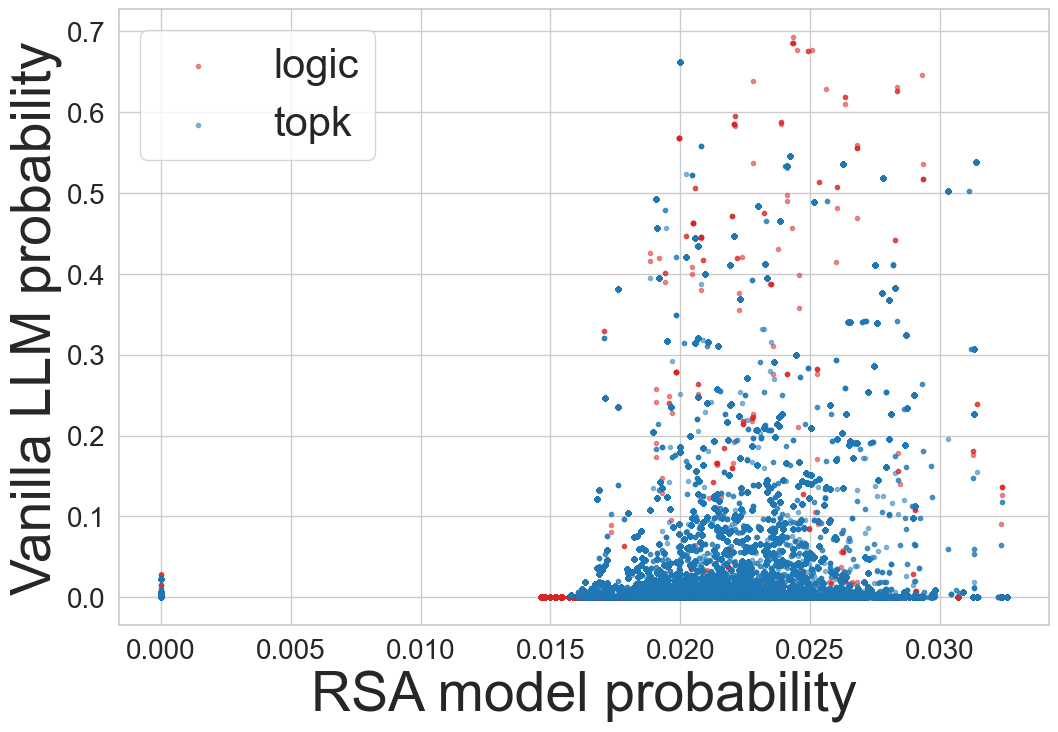}
            \caption{\pcc{}$_{topk}$ = 0.149, \pcc{}$_{logic}$ = 0.395, \pcc{}$_{all}$ = 0.288}
        \end{subfigure}
        \caption*{$\alpha = 0.2$}
        \label{fig:0.2}
    \end{subfigure}

    % Second pair: alpha = 0.6
    \begin{subfigure}{0.99\textwidth}
        \centering
        \begin{subfigure}{0.49\textwidth}
            \centering
            \includegraphics[width=\textwidth]{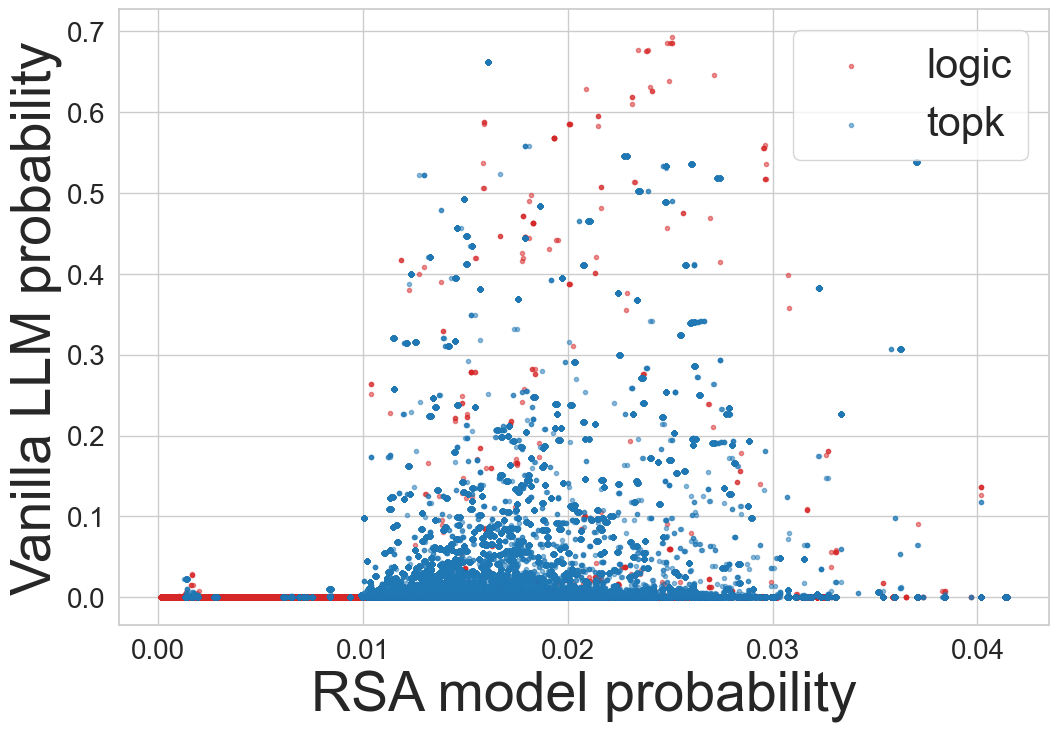}
            \caption{\pcc{}$_{topk}$ = 0.147, \pcc{}$_{logic}$ = 0.417, \pcc{}$_{all}$ = 0.288}
        \end{subfigure}
        \hfill
        \begin{subfigure}{0.49\textwidth}
            \centering
            \includegraphics[width=\textwidth]{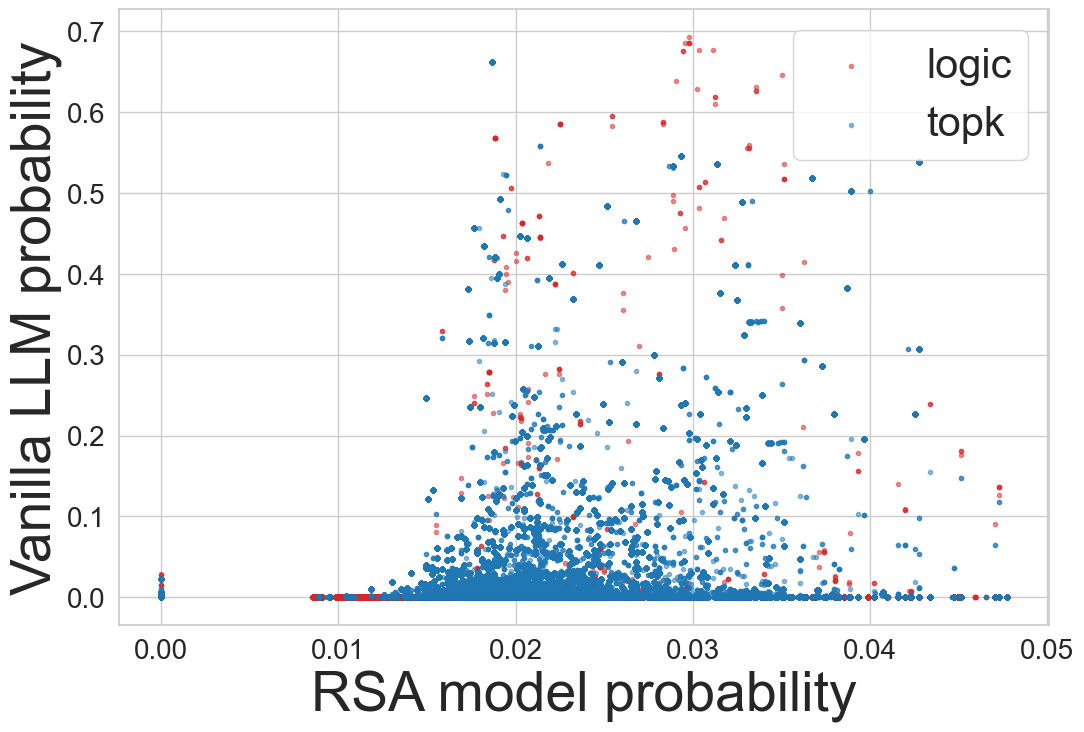}
            \caption{\pcc{}$_{topk}$ = 0.146, \pcc{}$_{logic}$ = 0.459, \pcc{}$_{all}$ = 0.304}
        \end{subfigure}
        \caption*{$\alpha = 0.6$}
        \label{fig:0.6}
    \end{subfigure}

    % Third pair: alpha = 1
    \begin{subfigure}{0.99\textwidth}
        \centering
        \begin{subfigure}{0.49\textwidth}
            \centering
            \includegraphics[width=\textwidth]{plots/a=1/all_prompt.png}
            \caption{\pcc{}$_{topk}$ = 0.139 , \pcc{}$_{logic}$ = 0.465, \pcc{}$_{all}$ = 0.291}
        \end{subfigure}
        \hfill
        \begin{subfigure}{0.49\textwidth}
            \centering
            \includegraphics[width=\textwidth]{plots/a=1/all_rule.png}
            \caption{\pcc{}$_{topk}$ =0.138 , \pcc{}$_{logic}$ =0.491, \pcc{}$_{all}$ = 0.303}
        \end{subfigure}
        \caption*{$\alpha = 1$}
        \label{fig:1}
    \end{subfigure}

    \caption{Results of the correlation scores of \llm{} and \rsa\ model using different $\alpha$ values. Each subplot group shows the overall correlation for scores in the \llm\ and \rsa\ models using (left) prompt-based MF and (right) rule-based MF. We report the \pcc{} scores for each utterances type.}
    \label{fig:alpha_comparison_part1}
\end{figure}

% Second part of the figure on the next page
\begin{figure}[!ht]
    \ContinuedFloat
    \centering

    % Fourth pair: alpha = 1.4
    \begin{subfigure}{0.99\textwidth}
        \centering
        \begin{subfigure}{0.49\textwidth}
            \centering
            \includegraphics[width=\textwidth]{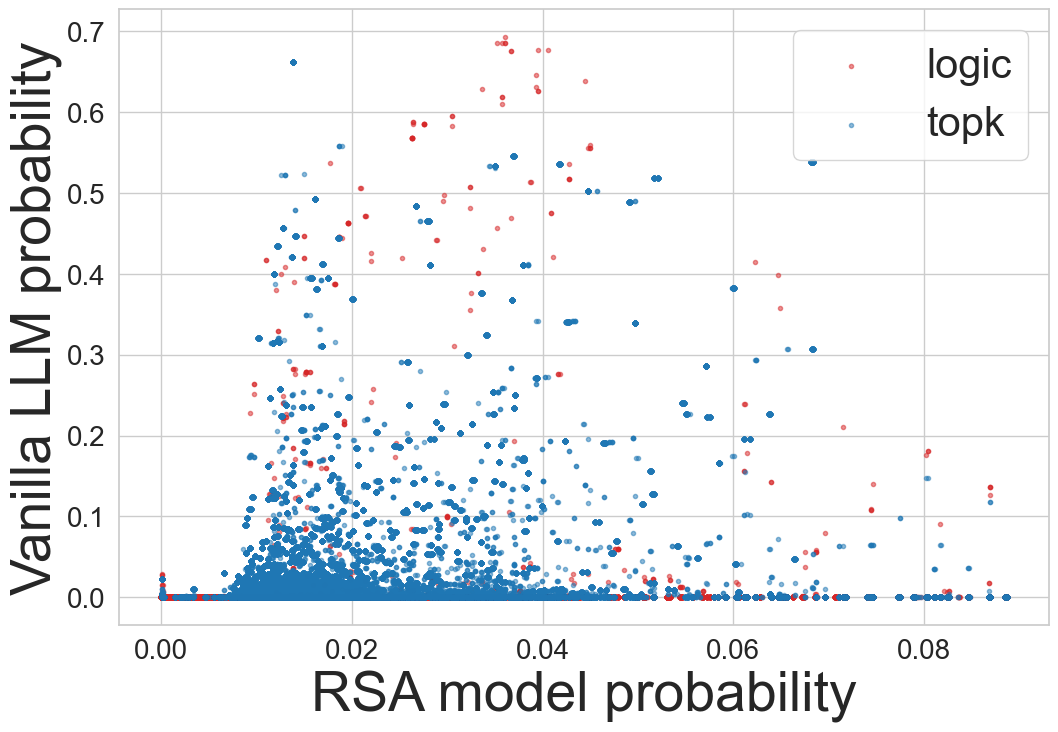}
            \caption{\pcc{}$_{topk}$ = 0.129, \pcc{}$_{logic}$ = 0.476, \pcc{}$_{all}$ = 0.278}
        \end{subfigure}
        \hfill
        \begin{subfigure}{0.49\textwidth}
            \centering
            \includegraphics[width=\textwidth]{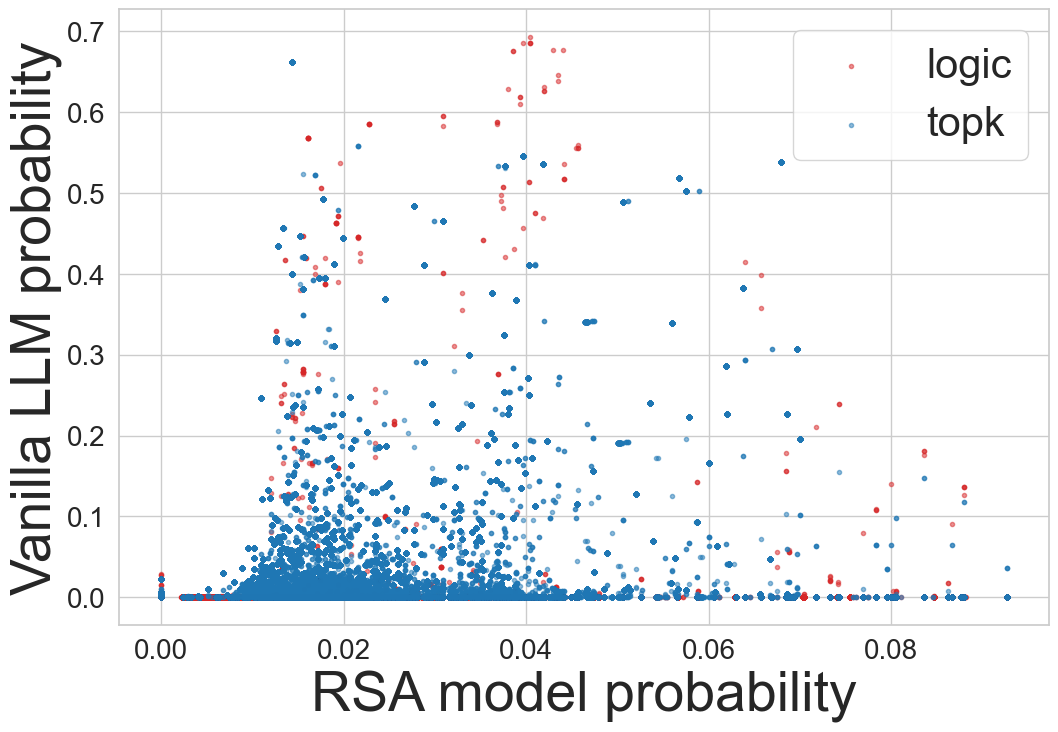}
            \caption{\pcc{}$_{topk}$ = 0.130, \pcc{}$_{logic}$ = 0.496, \pcc{}$_{all}$ = 0.290}
        \end{subfigure}
        \caption*{$\alpha = 1.4$}
        \label{fig:1.4}
    \end{subfigure}

    % Fifth pair: alpha = 1.8
    \begin{subfigure}{0.99\textwidth}
        \centering
        \begin{subfigure}{0.49\textwidth}
            \centering
            \includegraphics[width=\textwidth]{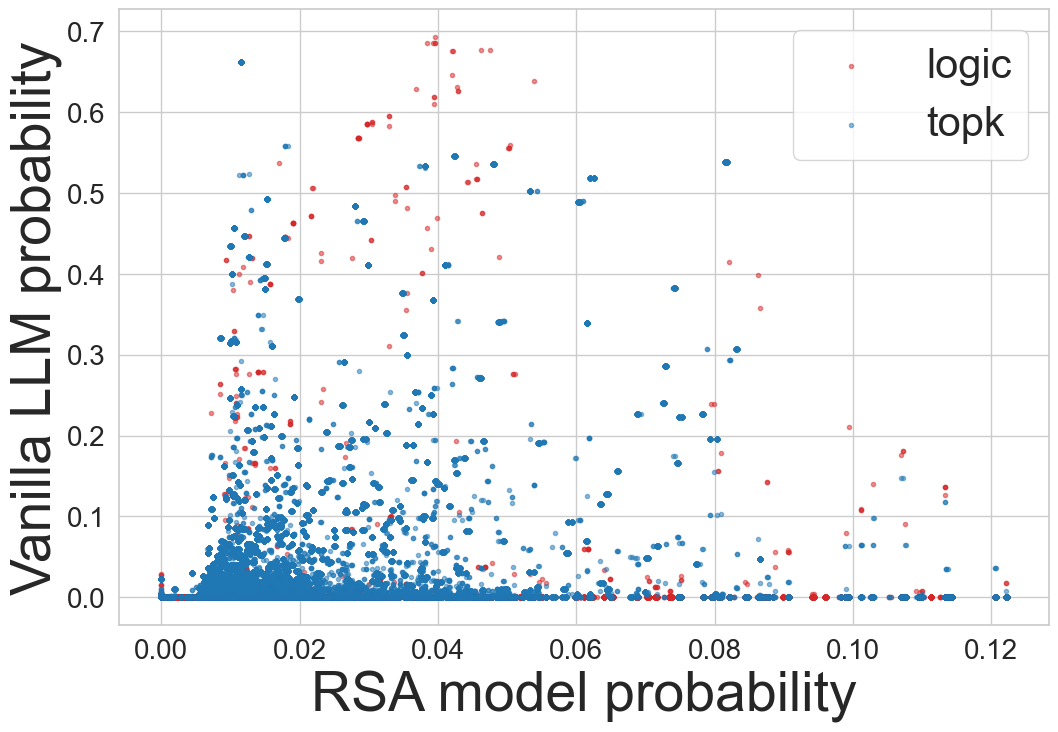}
            \caption{\pcc{}$_{topk}$ = 0.120,  \pcc{}$_{logic}$ = 0.465, \pcc{}$_{all}$ = 0.257}
        \end{subfigure}
        \hfill
        \begin{subfigure}{0.49\textwidth}
            \centering
            \includegraphics[width=\textwidth]{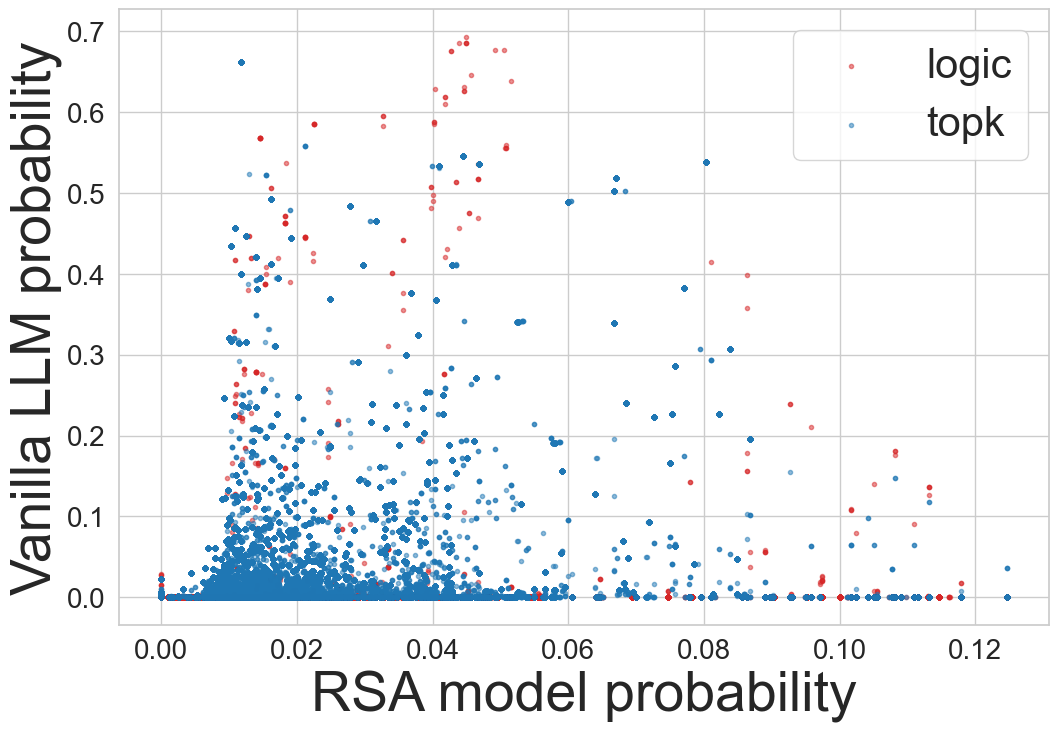}
            \caption{\pcc{}$_{topk}$ = 0.121,  \pcc{}$_{logic}$ = 0.482, \pcc{}$_{all}$ = 0.269}
        \end{subfigure}
        \caption*{$\alpha = 1.8$}
        \label{fig:1.8}
    \end{subfigure}

    % Sixth pair: alpha = 3
    \begin{subfigure}{0.99\textwidth}
        \centering
        \begin{subfigure}{0.49\textwidth}
            \centering
            \includegraphics[width=\textwidth]{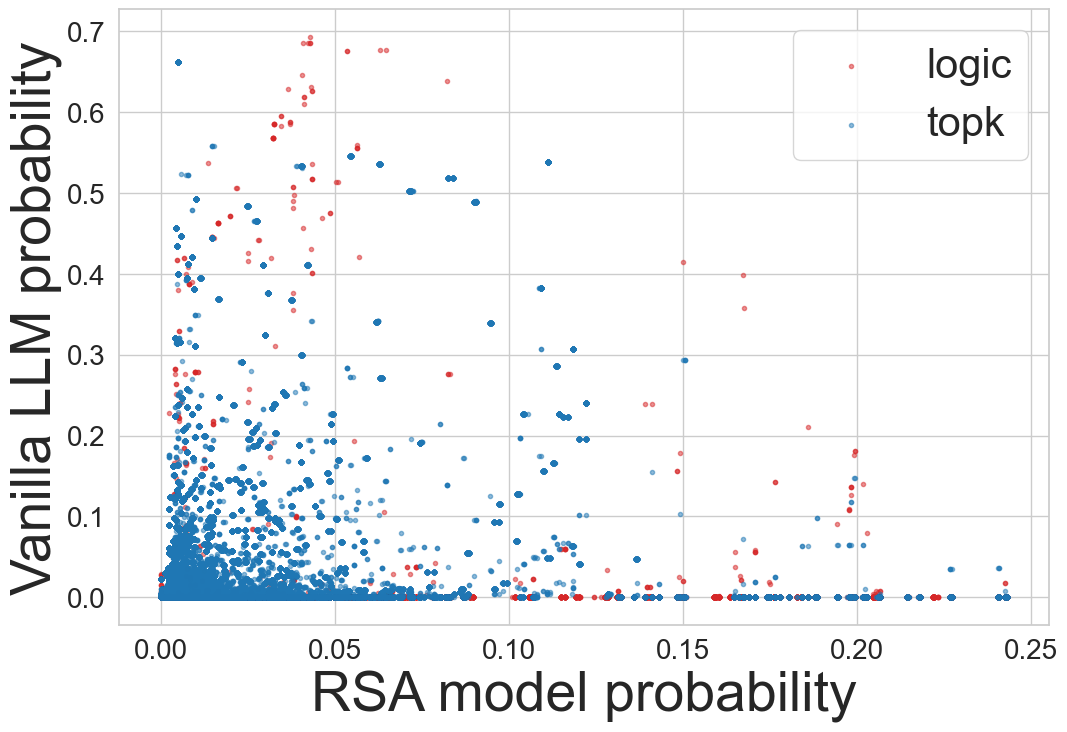}
            \caption{\pcc{}$_{topk}$ = 0.089,  \pcc{}$_{logic}$ = 0.394, \pcc{}$_{all}$ = 0.186}
        \end{subfigure}
        \hfill
        \begin{subfigure}{0.49\textwidth}
            \centering
            \includegraphics[width=\textwidth]{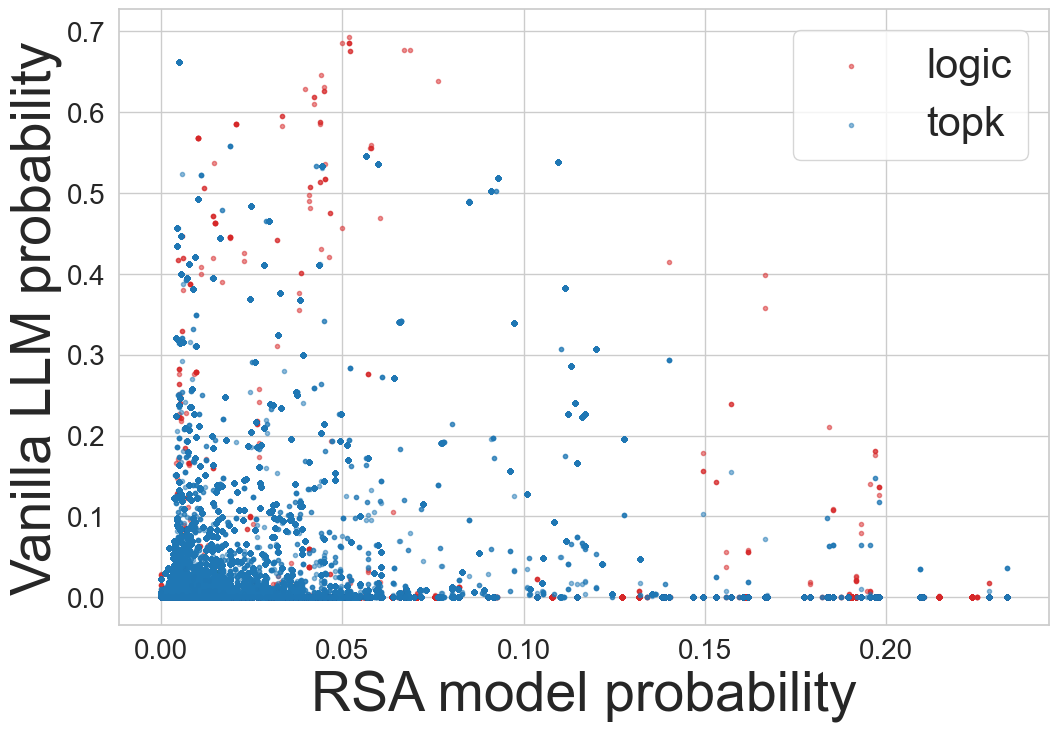}
            \caption{\pcc{}$_{topk}$ = 0.094,  \pcc{}$_{logic}$ = 0.412, \pcc{}$_{all}$ = 0.202}
        \end{subfigure}
        \caption*{$\alpha = 3$}
        \label{fig:3}
    \end{subfigure}

    \caption{Results of the correlation scores of \llm{} and \rsa\ model using different $\alpha$ values. Each subplot group shows the overall correlation for scores in the \llm\ and \rsa\ models using (left) prompt-based MF and (right) rule-based MF. We report the \pcc{} scores for each utterances type.}
    \label{fig:alpha_comparison_part2}
\end{figure}

\clearpage

\section{Meaning function evaluation} \label{app: mf_eval}
We evaluate the two proposed meaning functions (prompt-based MF and rule-based MF) by comparing their results on a set of test cases against human-labelled data. This evaluation is essential for fine-tuning parameters such as the number of examples given in the prompt, as well as for assessing the relative performance of the two meaning functions. We selected four constructed worlds for this purpose: ``topk1'' and ``topk2'', each comprising 2,056 $(o,u)$ pairs generated from top-k sequences, and ``logic1'' and ``logic2'', containing 497 and 602 pairs respectively, with utterances derived from rule-based logical constructions. 

\begin{figure}[!h]
    \centering
    \includegraphics[width=0.3\linewidth]{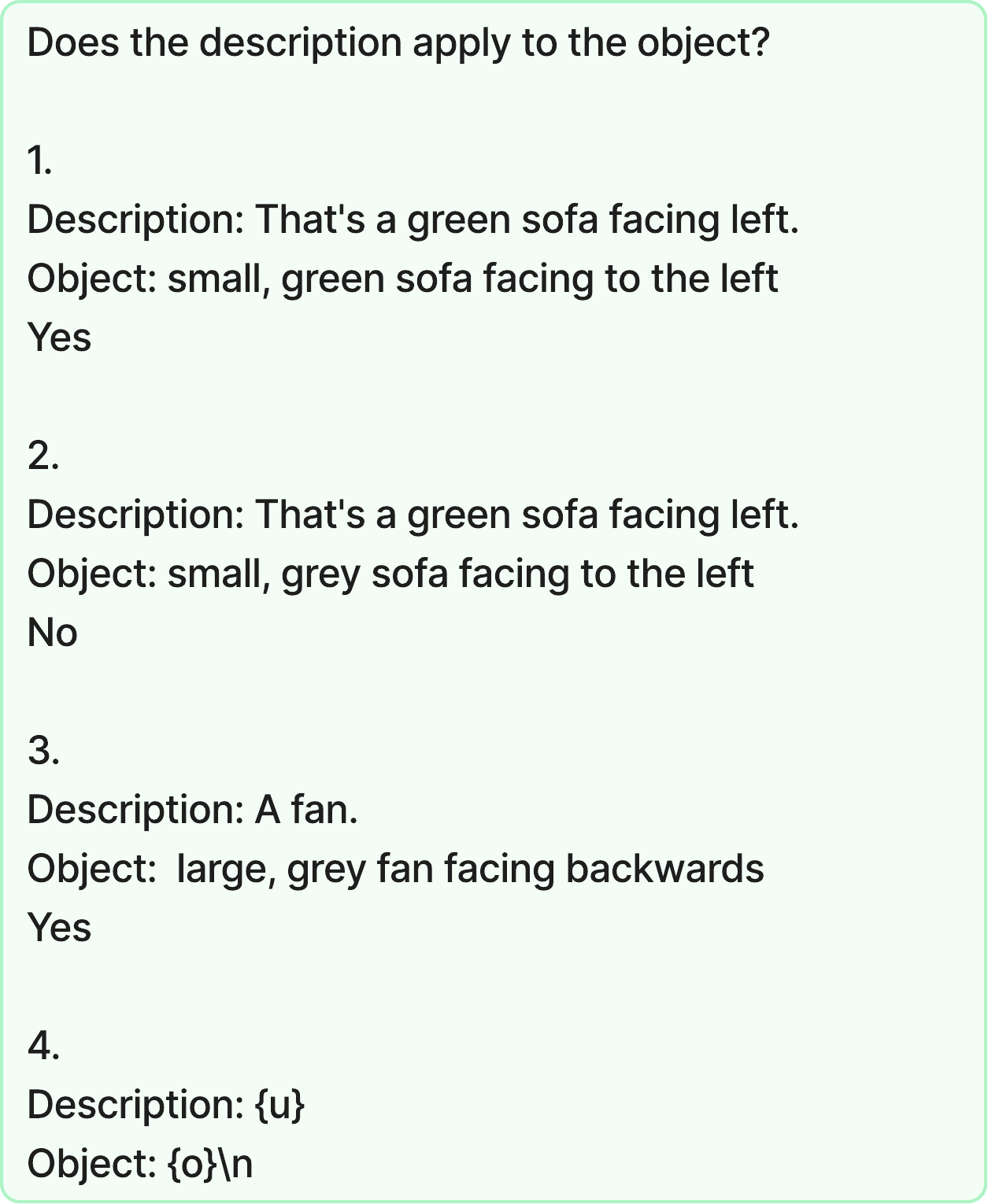}
    \caption{3-shot template for the prompt-based meaning function.}
    \label{fig:3-shot}
\end{figure}

\begin{figure}[!h]
    \centering
    \includegraphics[width=0.3\linewidth]{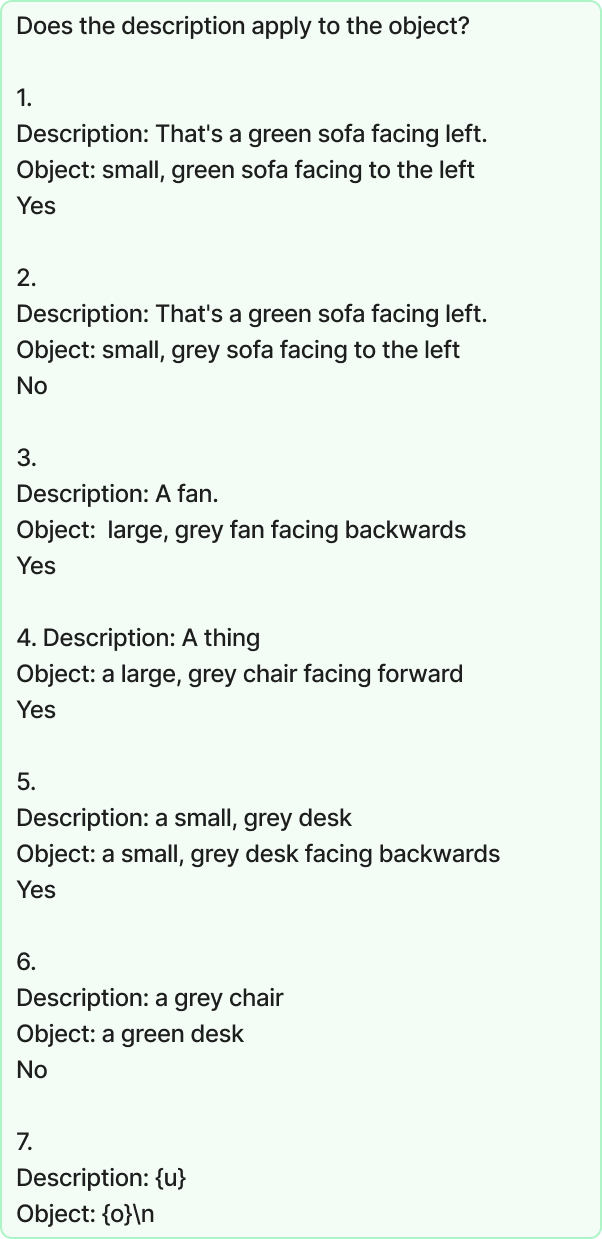}
    \caption{6-shot template for the prompt-based meaning function.}
    \label{fig:6-shot}
\end{figure}

For the prompt-based meaning function, we test with 3-shot prompts (Figure \ref{fig:3-shot}) and 6-shot prompts (Figure \ref{fig:6-shot}), and calculate the threshold $T$ that would give the best performance for each $n$-shot prompt setting. The threshold allows us to compare the prompt-based meaning function to our ground-truth annotations, by considering values of at least $T$ as $1$ and other values as $0$.

Table \ref{tab:prompt_meaning_function_eval} displays the performance of the prompt-based meaning function across different metrics using $n$-shot ($n = \{3, 6\}$) with optimised thresholds for best performance, as well as the performance of the rule-based meaning function. Overall, the rule-based meaning function consistently identifies the literal relationships of all logic-constructed pairs, but falls short on top-$k$ generated pairs that require natural language understanding ability for correct interpretation. For example, for the pair ($u_{\text{topk}} = $``the small grey chair that is not facing forwards'', $o = $``a small, grey chair facing backwards''), the rule-based meaning function scored it wrongly as it fails to interpret negation correctly.

The prompt-based method occasionally falls short as well. It generally performs better on the top-$k$ generated sequences comparatively to the logical constructed ones. Errors are more frequent when the sentence object is ``thing'' within the group of logically constructed sequences. Notably, the prompt-based meaning function achieves higher performance with our constructed 3-shot prompt compared to the 6-shot prompt.

We consider the rule-based meaning function particularly well-suited for our reference game setting due to the restricted attribute set for each object. Although the generated top-$k$ sequences may exhibit more diverse phrasings — such as when $u_{\text{topk}}$ is ``a tiny green table'' and $o$ is ``a small green desk'', where ``tiny'' and ``table'' are not present in the world vocabulary, the variations still fall within the attribute space of the furniture domain ($A_{\text{furniture}}$ = \{`Type', `Colour', `Orientation', `Size'\}). Consequently, the rule-based meaning function can effectively capture these variations through synonym mapping.

We anticipate that a more carefully crafted prompt or a more advanced language model could improve the performance of the prompt-based meaning function, although this would necessitate additional human and time resources. Nonetheless, this type of meaning function may be more suitable in a more flexible task setting, where the relationships between $o$ and $u$ extend beyond literal templates and require natural language understanding ability.

\begin{table}[!h]
    \centering
    \begin{tabular}{@{}r@{\qquad}ccclccclccc@{}}
    \toprule
        & \multicolumn{3}{c}{3-shot ($T$ = 0.6)} && \multicolumn{3}{c}{6-shot ($T$ = 0.8)} && \multicolumn{3}{c}{rule-based}\\
              & Acc & P & R && Acc & P & R && Acc & P & R \\
    \cmidrule(r){2-4}\cmidrule(lr){6-8}\cmidrule(l){10-12}
        % topk1 & 0.99 & 1.00 & 0.92  && 0.95 & 0.83 & 0.87 && 1.00 & 1.00 & 1.00 \\
        % topk2 & 0.98 & 0.96 & 0.89 && 0.97 & 0.98 & 0.83 && 1.00 & 1.00 & 1.00 \\
        % logic1 & 0.96 & 1.00 & 0.81 & 0.93 & 0.91 & 0.78 && 1.00 & 1.00 & 1.00 \\
        % logic2 & 0.95 & 1.00 & 0.71 & 0.94 & 0.91 & 0.78 && 1.00 & 1.00 & 1.00 \\
        topk1 & 0.997 & 0.992 & 0.988  && 0.991 & 0.973 & 0.976 && 0.999 & 0.996 & 0.996 \\ 
        topk2 & 0.987 & 0.925 & 1.000 && 0.972 & 0.875 & 0.966 && 0.999 & 1.000 & 0.996\\
        logic1 & 0.956 & 1.000 & 0.804 && 0.966 & 0.970 & 0.875 && 1.000 & 1.000 & 1.000 \\
        logic2 & 0.952 & 0.966 & 0.768 && 0.938 & 0.830 & 0.830 && 1.000 & 1.000 & 1.000 \\
    \bottomrule
    \end{tabular}
    \caption{Performance of the prompt-based and rule-based meaning function across different metrics (Accuracy, Precision and Recall), the prompt-based meaning functions are using $n$-shot ($n = \{3, 6\}$) with optimised thresholds $T$ for best performance.}
    \label{tab:prompt_meaning_function_eval}
\end{table}

\clearpage
\section{Experiment 1: Evaluation between each reference game}\label{app: experiment1}
Table \ref{tab:meaning_result_each_refgame} presents the mean scores and standard deviations of \pcc\ and \srcc\ across all reference games, when the utterance space is composed of the two different utterance types, and the \rsa\ model is calculated with two different meaning functions. 

Overall, the six model evaluations all show a positive correlation between the scoring of the vanilla \llm{} and the \rsa\ model. The results indicate a preference for utterance type in the meaning function. The \pcc\ and \srcc\ scores for logic-constructed utterances are higher when the \rsa\ model is configured with a rule-based meaning function. However, the \llm{} shows stronger alignment with the top-$k$ utterances compared to the \rsa\ configured with a prompt-based meaning function. This aligns with our evaluation of the meaning function: the rule-based meaning function achieves full accuracy when judging pairs in a controlled setting but falls short in flexible settings requiring advanced natural language understanding for interpretation.

Focusing on the \srcc\ scores, an interesting observation is that while the \srcc\ score for top-$k$ utterances is very low, the \srcc\ score for all utterances, regardless of type, is high (0.606 and 0.736, respectively). This suggests that the \llm{} can rank sequences of different constructed types correctly to some extent, but lacks the advanced pragmatic ability to rank pragmatic sequences accurately, as indicated by the low \srcc\ scores for the top-$k$ cases (0.086 and 0.059, respectively).

% In terms of the utterance type, when we only investigate the correlation when the models are scoring the logic-constructed sequences, the correlation of the models' scoring are more strongly correlated than that of the top-$k$ constructed sequences. In terms of different types of \rsa\ models, the table suggests that the vanilla \llm{} has a better correlation to the \rsa\ model that is calculated with a rule-based meaning function.

\begin{table}[!h]
    \centering
    \begin{tabular}{@{}clcccc@{}}
    \toprule
    && \multicolumn{2}{c}{\textbf{PCC}} & \multicolumn{2}{c}{\textbf{SRCC}}  \\
    \cmidrule(r){3-4} \cmidrule(l){5-6}
    \textbf{Utt. Type}& \textbf{RSA MF} & \textbf{Mean} & \textbf{$\sigma$} & \textbf{Mean} & \textbf{$\sigma$} \\
    \midrule
    % Logic
    \multicolumn{1}{c}{\multirow{2}{*}{Logic}}&Prompt-based & 0.465 & 0.118 & 0.253 & 0.128 \\
    &Rule-based & \textbf{0.491} & 0.114 & \textbf{0.460} & 0.051 \\
    \midrule

    % Top-k
    \multicolumn{1}{c}{\multirow{2}{*}{Top-k}}&Prompt-based & \textbf{0.139} & 0.145 & \textbf{0.086} & 0.173 \\
    &Rule-based & 0.138 & 0.144 & 0.059 & 0.163   \\ \midrule

    % All
    \multicolumn{1}{c}{\multirow{2}{*}{All}}&Prompt-based & 0.291 & 0.082 & 0.606 & 0.103\\
    &Rule-based &\textbf{ 0.303} & 0.083 & \textbf{0.736} & 0.078\\
    \bottomrule
    
    \end{tabular}
    \caption{Comparison of the mean and standard deviation ($\sigma$) for PCC and SRCC metrics across different reference games, highlighting the correlation between the \llm{} model and two \rsa\ models using different meaning functions (MFs), and with different utterance types.}
    \label{tab:meaning_result_each_refgame}
\end{table}

\end{document}